\documentclass[12pt, a4paper]{article}

\usepackage[utf8]{inputenc}
\usepackage[T1]{fontenc}
\usepackage{lmodern}
\usepackage[margin=1in]{geometry}
\usepackage{setspace}
\usepackage{hyperref}
\usepackage{url}
\usepackage{graphicx}
\usepackage{amsmath}
\usepackage{amssymb}
\usepackage{booktabs}
\usepackage{array}
\usepackage{longtable}
\usepackage{multirow}
\usepackage{xcolor}
\usepackage[numbers]{natbib}
\usepackage{caption}
\usepackage{subcaption}
\usepackage{float}
\usepackage{microtype}
\usepackage{parskip}
\usepackage{titlesec}
\usepackage{fancyhdr}
\usepackage{abstract}
\usepackage{authblk}

\titleformat{\section}{\large\bfseries}{\thesection.}{0.5em}{}
\titleformat{\subsection}{\normalsize\bfseries}{\thesubsection.}{0.5em}{}
\titleformat{\subsubsection}{\normalsize\itshape}{\thesubsubsection.}{0.5em}{}

\hypersetup{
    colorlinks=true,
    linkcolor=black,
    citecolor=black,
    urlcolor=blue
}

\begin{document}

\title{
    \LARGE\textbf{Can Machine Learning Forecast Rice Yields in
    Data-Constrained Settings? Satellite Climate Data, National Crop
    Statistics, and Lessons from Sierra Leone}\\[1em]
    \normalsize\textit{Evidence for the Feed Salone Strategy 2023--2030}
}

\author[1,2]{Ibrahim Denis Fofanah}

\affil[1]{
    Seidenberg School of Computer Science \& Information Systems\\
    Pace University, New York, USA
}
\affil[2]{
    RiseAfrica Foundation for STEM and Innovation\\
    Sierra Leone, West Africa
}

\maketitle
\thispagestyle{empty}

\begin{abstract}
\noindent
Agriculture employs the majority of Sierra Leone's rural population, yet the
sector operates with almost no data-driven decision support, and no published
machine learning study has examined Sierra Leonean crop yields specifically.
This paper provides the first such evidence base, and asks a question with
direct consequences for the country's Feed Salone Strategy (2023--2030): can
rice yield be forecast from the data Sierra Leone currently has?

Using 25 years of FAOSTAT production data (2000--2024) for nine major crops,
we train three ensemble algorithms (XGBoost, Gradient Boosting, Random Forest)
under a strict anti-leakage protocol: lagged predictors only, and an
expanding-window walk-forward evaluation across seven held-out years
(2018--2024), benchmarked against naive persistence. The answer is no:
no model trained on crop statistics alone outperforms simply carrying forward
the previous year's yield.

We then augment the models with freely available satellite climate data:
CHIRPS rainfall and NASA POWER temperature, aggregated to national
growing-season features. This reverses the result. A climate-only XGBoost
model reduces forecast error by a third relative to persistence (RMSE 284 vs.\
428 kg/ha), a gain that holds for a linear model as well and is robust to
excluding the anomalous 2018 season. Early-season (May--June) rainfall is the
dominant predictor, implying that seasonal yield risk is observable months
before harvest. Two boundaries are documented. No model anticipated the 2018
yield collapse, whose origins were institutional rather than climatic, and
the 2020--2022 record yields occurred despite below-average rainfall,
consistent with input-driven gains under recent policy programs.

The findings carry a clear message for Feed Salone: Sierra Leone's existing
agricultural statistics cannot support yield forecasting, but combining them
with free satellite climate data already can, and household-level microdata
would extend prediction to the district level where decisions are made. The
full pipeline is open-source and replicable for other data-constrained
agricultural economies.

\end{abstract}

\newpage

\section{Introduction}

\subsection{Agriculture, Data, and the Intelligence Gap}

In the twenty-first century, agriculture has become as much an information
problem as an agronomic one. Across the world's most productive farming
systems, data science has changed how farmers make decisions. Satellite imagery
informs planting schedules. Machine learning models forecast seasonal yields
before a single seed enters the ground. Real-time price information tells
producers where to sell and when. The result is a farming sector that is not
merely productive but intelligent: responsive to information, adaptive to risk,
and capable of generating wealth at scale.

Sub-Saharan Africa has not shared equally in this transformation. Smallholder
farmers, who operate plots typically smaller than two hectares and account for
the majority of food production across the region, continue to make
consequential agricultural decisions without the information those decisions
require \citep{worldbank2022}. What to plant this season? When will the rains
arrive? What price will the market offer at harvest? These are not abstract
questions. They determine whether a farming household eats well or faces
hunger, and they are answered not with models or forecasts but with memory,
tradition, and approximation. The consequences of that gap fall most heavily
on those least equipped to absorb them.

This paper is concerned with one country where that gap is particularly acute,
and where the stakes of closing it are particularly high: Sierra Leone.

\subsection{Sierra Leone: Agricultural Potential and Persistent Underperformance}

Sierra Leone possesses an estimated 5.4 million hectares of fertile arable
land, a tropical climate that supports year-round cultivation, and river
systems that feed inland valley swamps across every region \citep{usda2023}.
Rice, the country's staple food, grows naturally and abundantly across both
lowland and upland ecologies. Cassava, sweet potato, groundnut, oil palm,
cocoa, and coffee round out an agricultural portfolio that could, in principle,
feed the nation and generate substantial export revenue.

In practice, the sector has chronically underperformed. Sierra Leone produced
approximately 1.4 billion kilograms of rice in 2023, yet this was insufficient
to meet national demand \citep{statssl2023}. The country imported 480,000
metric tons of rice in 2022 alone, a dependency that drains foreign exchange
reserves and leaves food security vulnerable to global price shocks
\citep{usda2023}. More troublingly, 75 percent of arable land remains
uncultivated \citep{usda2023}, which points not to a scarcity of land or
labor but to a scarcity of the enabling conditions that make cultivation
worthwhile: infrastructure, inputs, market access, and decision support.

A further dimension of the problem, which motivates the broader research
agenda of which this paper is the first step, is post-harvest loss.
Smallholder farmers across sub-Saharan Africa lose between 30 and 50 percent
of their harvest to spoilage, poor storage, inadequate processing, and
transport failures \citep{affognon2015, fao2019}. In Sierra Leone, threshing
and winnowing are performed by hand, post-harvest drying occurs on mud floors
and tarmac roads, and access to even basic concrete drying infrastructure is
limited to a small share of farming communities \citep{nrds2022}. The economic
loss is substantial, but the deeper consequence is food insecurity: as of
2023, 39 percent of Sierra Leone's population lived below the poverty line
\citep{worldbank2023}.

\subsection{The Policy Context: Feed Salone}

In October 2023, President Julius Maada Bio launched the Feed Salone Strategy
as the flagship initiative of the Medium-Term National Development Plan
2024--2030, with a mandate to transform Sierra Leone's food systems from
subsistence and dependency toward a resilient, commercially oriented, and
technology-driven sector \citep{feedsalone2023}. The strategy targets, among
other goals, a doubling of rice production, a reduction in post-harvest losses,
and the integration of digital and data-driven tools into agricultural
extension and planning \citep{feedsalone2023}.

Yet the Ministry of Agriculture and Food Security's own assessment identifies a
critical constraint: MAFS lacks the data infrastructure and technical capacity
needed to monitor, evaluate, and guide its own policies \citep{fidinnovation2024}.
Field staff in rural areas lack training in even basic analytical tools. The
Planning, Evaluation, Monitoring, and Statistics Division manages a
decentralized monitoring system that struggles to produce timely,
district-level intelligence \citep{fidinnovation2024}. The ambition of Feed
Salone runs ahead of the data systems available to support it.

This paper responds to that gap, not by assuming that data-driven forecasting
is possible in Sierra Leone, but by testing rigorously whether it is, with
which data, and under what limits.

\subsection{The Research Gap}

Machine learning applications in agricultural prediction have grown
substantially over the past decade. Ensemble methods, particularly Random
Forest, XGBoost, and Gradient Boosting, have consistently demonstrated strong
predictive performance for crop yield estimation across diverse agricultural
environments \citep{vanklompenburg2020, liakos2018}. Explainable AI frameworks
have made these models increasingly useful to non-specialist users
\citep{arrieta2020}.

Yet the geographic distribution of this work is skewed. Systematic reviews of
the crop yield prediction literature find that while South Africa, Ghana, and
East Africa are increasingly well represented, West Africa, and Sierra Leone
specifically, remains virtually absent from the published literature
\citep{frontiers2026}. The only published study directly examining
post-harvest losses in Sierra Leone used traditional statistical methods on a
sample of 232 rice farmers across eight districts \citep{tandason2023}. No
machine learning study has been built specifically on Sierra Leonean yield
data.

There is a second, methodological gap that this paper addresses directly.
Much of the small-sample agricultural ML literature reports very high
in-sample or randomly cross-validated accuracy ($R^2 > 0.95$ is common) on
national annual time series of only a few dozen observations. As
Section~\ref{sec:validation} discusses, such designs are highly vulnerable to
target leakage and to temporal information bleeding across random
train--test splits \citep{kaufman2012, bergmeir2012}. This study adopts a
strict anti-leakage protocol and a walk-forward evaluation, and benchmarks
every model against naive baselines, a discipline that, as the results show,
changes the conclusions entirely.

\subsection{Research Questions}

Three research questions guide this study:

\begin{enumerate}
    \item Can rice yield in Sierra Leone be forecast from the country's
    existing agricultural statistics alone, using 25 years of FAOSTAT crop
    data (2000--2024), when evaluated against naive baselines under
    walk-forward validation?

    \item Does augmenting those statistics with freely available satellite
    climate data (CHIRPS rainfall, NASA POWER temperature) change the answer,
    and which climate signals carry the predictive weight?

    \item What do the results imply for the implementation of the Feed Salone
    Strategy 2023--2030, and for the data-infrastructure investments needed to
    make data-driven agricultural governance sustainable in Sierra Leone?
\end{enumerate}

\subsection{Contributions}

Four contributions emerge from this work. First, it provides the first
machine learning evidence base built specifically for Sierra Leonean
agriculture, establishing under rigorous evaluation what the country's
existing public data can and cannot support. Second, it demonstrates that
free satellite climate data, requiring no accounts, licenses, or fees, is
sufficient to move national rice yield forecasting from impossible to useful,
reducing out-of-sample error by one third relative to naive persistence.
Third, it documents, as a methodological case study, how standard but flawed
designs (same-year features, random train--test splits) produce illusory
accuracy on exactly this kind of small national time series. Fourth, it
translates the findings into concrete, data-grounded policy recommendations
aligned with the Feed Salone Strategy, and archives the full open-source
pipeline for replication in other data-constrained agricultural economies.

\subsection{Paper Organization}

Section 2 reviews the relevant literature. Section 3 describes the study
context and the three data sources. Section 4 presents the methodological
framework, including the anti-leakage protocol and validation design.
Section 5 reports results. Section 6 discusses findings, limits, and policy
implications. Section 7 concludes with policy recommendations.

\section{Literature Review}

\subsection{Machine Learning in Agricultural Prediction}

The application of machine learning to crop yield prediction has matured
considerably over the past decade. Early approaches relied on linear
regression and simple statistical models that estimated yield as a function of
rainfall and temperature \citep{lobell2010}. The limitations of those
approaches, chiefly their inability to capture non-linear relationships,
interaction effects, and spatial variation, drove a shift toward ensemble
methods and deep learning architectures that now dominate the field.

Among supervised learning algorithms, Random Forest, XGBoost, and Gradient
Boosting Machines have emerged as the most consistently reliable for crop
yield prediction across diverse agricultural environments
\citep{vanklompenburg2020, liakos2018}. Deep learning architectures,
particularly LSTM networks and CNNs, have shown strong results when
integrating satellite-derived vegetation indices and remote sensing data
\citep{frontiers2026}. However, these architectures demand large training
datasets and computational resources that are rarely available in
data-constrained settings. For smallholder agriculture in sub-Saharan Africa,
where short administrative time series rather than dense satellite stacks
constitute the primary structured data source, ensemble methods offer a more
practical and interpretable path to prediction \citep{arrieta2020}.

The interpretability question deserves emphasis. A recurring critique in the
agricultural machine learning literature is that black-box models, however
accurate, are of limited use to farmers, extension officers, and policymakers
who need to understand and trust what a model is telling them
\citep{frontiers2026}. SHAP (SHapley Additive exPlanations) has largely
addressed this by providing feature-level attribution that makes model
outputs readable to non-specialist users \citep{lundberg2017, arrieta2020}.
The integration of SHAP analysis into this study reflects that priority.

\subsection{Validation Pitfalls in Small-Sample Yield Prediction}
\label{sec:litvalidation}

A growing methodological literature cautions that predictive accuracy claims
in time-series settings are only as credible as the validation design behind
them. Two failure modes are especially relevant to national annual yield
series. The first is \textit{target leakage}: the inclusion of predictors
that are functions of, or near-duplicates of, the outcome variable
\citep{kaufman2012}. In national crop statistics this arises easily.
An aggregate ``cereals'' indicator in a country where one cereal dominates
production, for example, is arithmetically almost identical to that crop's
own series. The second is \textit{temporal leakage through random splitting}:
when annual observations are assigned randomly to train and test sets,
test-year information enters training through lagged and rolling features of
neighboring years, and the evaluation no longer simulates genuine forecasting
\citep{bergmeir2012, roberts2017}. Both failure modes inflate $R^2$
dramatically on small samples. The appropriate remedies, namely strictly
lagged or exogenous predictors, expanding-window walk-forward evaluation, and
comparison against naive baselines, are adopted throughout this study and
described in Section~\ref{sec:validation}.

\subsection{Machine Learning Applications in Sub-Saharan African Agriculture}

Research applying machine learning to agricultural prediction in sub-Saharan
Africa has grown meaningfully in recent years, though it remains concentrated
in a handful of countries: Kenya, Rwanda, Nigeria, South Africa, and Ethiopia
account for the majority of published work, while West Africa, and the
smaller economies of the Mano River Union in particular, is conspicuously
underrepresented \citep{frontiers2026}.

A country-level study on West African crop yield prediction trained models on
FAOSTAT data from 1990 to 2020 across fourteen countries, including Sierra
Leone \citep{sciencedirect2022}. While that work showed that ML-based yield
prediction is feasible in the West African context, its multi-country pooling
and national-level aggregation obscure the heterogeneity that matters most
for agricultural policy. A farmer in Kailahun District and a farmer in Port
Loko District face different soil conditions, rainfall patterns, and market
access constraints; a national average prediction is not directly useful to
either of them. This study responds by developing a Sierra Leone-specific
analytical pipeline calibrated to local data and policy context, while being
explicit, in its limitations, that national aggregation remains a binding
constraint that only household- and district-level microdata can relax.

\subsection{Post-Harvest Loss Research in West Africa}

Post-harvest losses represent one of the most significant and least addressed
threats to food security in sub-Saharan Africa. Meta-analyses across the
region estimate that smallholder farmers lose between 30 and 50 percent of
their production annually to physical spoilage, quality degradation, and
market failures \citep{affognon2015}. The causes are well documented:
inadequate storage infrastructure, poor temperature and humidity management,
delayed threshing and drying, pest and fungal contamination, and distance
from processing facilities consistently appear as primary contributors
\citep{sheahan2017, fao2019}. Socioeconomic factors, including household
poverty, limited credit access, and gender dynamics in post-harvest labor,
add further complexity \citep{tandason2023}.

What is missing is the application of machine learning to post-harvest loss
\textit{prediction}: models that could identify high-risk communities before
losses happen, enabling proactive rather than reactive intervention. Such
models require household-level data on storage, transport, and handling
practices. These data exist in Sierra Leone's 2023 national agricultural
survey but are not yet accessible (Section~\ref{sec:data}). Post-harvest loss
modeling is therefore positioned in this paper as the next step of the
research agenda rather than a delivered result.

\subsection{The Sierra Leone Agricultural Data Landscape}

Sierra Leone's agricultural data environment has improved in recent years,
driven largely by the 50x2030 Initiative, a partnership between FAO, IFAD, the
World Bank, and national statistical agencies aimed at transforming
agricultural data systems in low-income countries \citep{50x2030_2024}. The
2023 Sierra Leone Annual Agricultural Sample Survey (SLAASS), conducted by
Statistics Sierra Leone in partnership with MAFS, represents the most
comprehensive nationally representative agricultural household survey in the
country's history, covering 5,200 households across 520 enumeration areas
\citep{statssl2023}.

Despite this progress, important structural challenges remain. MAFS's
Planning, Evaluation, Monitoring, and Statistics Division manages a
decentralized data collection system that faces real limitations in rural
areas, where field staff lack training in analytical tools
\citep{fidinnovation2024}. A Data Ecosystem Mapping published in May 2024
under the 50x2030 Initiative concluded that a large gap between data
collection and data use for decision-making persists \citep{50x2030_2024}.

This paper engages directly with that gap. By establishing empirically what
existing publicly available data can and cannot support, it replaces a
rhetorical case for data-infrastructure investment with a measured one.

\section{Study Context and Data}
\label{sec:data}

\subsection{Agricultural Profile of Sierra Leone}

Sierra Leone is a small coastal nation in West Africa with a population of
approximately 8.4 million people, of whom roughly 60 percent live in rural
areas and depend on agriculture for their livelihoods \citep{worldbank2023}.
Agriculture contributes approximately 25 percent of GDP as of 2024
\citep{wdi2024}, a figure that understates the sector's importance to rural
welfare, given that agricultural income is the primary livelihood source for
the majority of the country's poor.

The agricultural system is dominated by smallholders. According to the 2023
SLAASS, 92.4 percent of agricultural households have between one and three
economically active members, with most operating plots smaller than two
hectares \citep{statssl2023}. Rice is the dominant staple crop, produced
across all five administrative regions. Sweet potato, cassava, groundnut, and
oil palm are also widely cultivated. The Eastern Region holds the largest
share of agricultural plots at 31.8 percent, followed by the Southern Region
at 20.4 percent and the North-Western Region at 19.6 percent
\citep{statssl2023}.

\subsection{Crop Statistics: FAOSTAT, 2000--2024}

The first data source is the FAOSTAT Crops and Livestock Products database,
maintained by the Food and Agriculture Organization of the United Nations
\citep{fao2019}. FAOSTAT provides standardized, nationally aggregated
agricultural statistics compiled from official government submissions, FAO
estimates, and imputation procedures designed to ensure time-series
consistency. The database is freely accessible and requires no approval to
download, which makes it a practical and fully reproducible foundation for
research in data-constrained settings.

For Sierra Leone, the data cover the years 2000 through 2024, yielding a
25-year annual time series. Three indicators are extracted for each crop:
area harvested (hectares), yield (kg/ha), and production quantity (tonnes).
Nine crops with complete 25-year coverage on all three indicators are
included: rice, cassava, maize, groundnuts, oil palm fruit, sweet potato,
sorghum, cocoa beans, and plantains. Yams, although agriculturally relevant,
are reported in FAOSTAT for Sierra Leone only from 2022 onward (3 of 25
years) and are therefore excluded, as are FAOSTAT aggregate items (e.g.,
``Cereals, primary'') and processed products, for reasons developed in
Section~\ref{sec:features}. The raw dataset and all processing code are
archived in the study's public GitHub repository
(\url{https://github.com/Denis060/sierraleone-agri-ml}) to ensure full
reproducibility.

Two properties of this source deserve explicit acknowledgment. First, FAOSTAT
provides national-level aggregates only; it contains no household, plot, or
district observations, and therefore cannot support the post-harvest loss
modeling component of this research agenda, which requires microdata on
storage, transport, and handling. The 2023 SLAASS would provide that
microdata but was not accessible during the study period despite registration
on the FAO microdata portal; it is treated as the primary direction for
future research (Section~\ref{sec:limitations}). Second, a substantial share
of the FAOSTAT values for Sierra Leone are not official national figures:
across the full extract, only 467 of 2,593 records carry FAO's ``official
figure'' flag, with the remainder estimated or imputed by FAO or receiving
agencies. For the rice yield outcome series specifically, 14 of 25 annual
values are flagged official, and the most recent years (2022--2024) are FAO
estimates. Some of the smoothness that any model exploits in this series
therefore originates in FAO's imputation procedures rather than in the
underlying agricultural system, which is itself evidence of the thinness of
the national statistical record that this paper's policy recommendations
address.

\subsection{Satellite Climate Data: CHIRPS and NASA POWER}
\label{sec:climatedata}

The second and third data sources are free, globally available satellite
climate products requiring no registration or license.

\textbf{CHIRPS rainfall.} Monthly precipitation for 2000--2024 is taken from
the Climate Hazards Group InfraRed Precipitation with Station data (CHIRPS
v2.0), a quasi-global 0.05$^{\circ}$ gridded product blending satellite
imagery with in-situ station data \citep{funk2015}. Monthly Africa rasters
are spatially averaged over Sierra Leone's national boundary (geoBoundaries
ADM0 \citep{runfola2020}) to produce a national monthly rainfall series.

\textbf{NASA POWER temperature.} Monthly 2-meter air temperature (T2M) for
2000--2024 is taken from the NASA POWER agroclimatology API, sampled on a
five-point grid spanning the country (approximately 7.0--9.9$^{\circ}$N,
10.3--13.3$^{\circ}$W) and averaged into a national series. NASA POWER's
precipitation product is retained solely as an independent cross-check on
CHIRPS; the two monthly rainfall series correlate at $r = 0.73$, providing
reassurance that the national rainfall signal is not an artifact of a single
product.

\subsection{Variable Construction}
\label{sec:features}

\subsubsection{Outcome Variable: Rice Yield}

The outcome variable is rice yield in kilograms per hectare, extracted from
the FAOSTAT yield element for Sierra Leone across the full 25-year study
period. The resulting annual series provides a continuous outcome variable
suitable for supervised regression.

\subsubsection{Predictor Variables and the Anti-Leakage Protocol}

Predictors are organized into three groups, governed by one rule: \textbf{no
predictor may contain same-year information about agricultural outcomes.}
Every crop-derived feature must be knowable before the prediction year's
harvest. Exogenous climate observations are exempt from the lag requirement
because they are measured independently of the outcome and are available
during the growing season, before yield is realized.

\textbf{Group 1: lagged crop indicators (35 features).} One-year lags of
area harvested, yield, and production for the nine crops, excluding rice's
own contemporaneous values. Three classes of variables used in earlier,
flawed versions of this pipeline are excluded by construction: (i) rice
production per hectare, which is arithmetically the target; (ii) FAOSTAT
aggregates such as ``Cereals, primary,'' of which rice constitutes 87--93
percent of production in every year of the series (correlation with rice
yield: $r = 0.998$), making them near-duplicates of the outcome; and (iii)
all same-year values of any crop variable.

\textbf{Group 2: autoregressive features.} Rice yield at lags of one, two,
and three years, plus three-year and five-year rolling means computed
strictly over past years (the series is shifted before the rolling window is
applied, so the window for year $t$ never touches year $t$).

\textbf{Group 3: climate features (used in the climate configurations).}
Growing-season total rainfall (May--October), early-season rainfall
(May--June), peak-season rainfall (July--September), a standardized
growing-season rainfall anomaly whose climatology is computed only from
pre-test-window years (2000--2017), the maximum one-month rainfall deficit
within the growing season, mean growing-season temperature, and prior-year
growing-season rainfall. Structural indicators (linear year trend; Ebola
2014--2016, COVID-19 2020--2021, and Feed Salone 2023+ dummies) complete the
set.

A runtime assertion in the published pipeline fails the build if any
same-year crop variable reaches the predictor matrix.

\section{Methodology}

\subsection{Research Design}

Rice yield forecasting is treated as a supervised regression problem with an
explicit forecasting simulation: in every evaluation, the model may use only
information that would have been available before the year being predicted.
Three feature configurations are compared (lagged crop statistics only,
climate features only, and their combination) across three ensemble learners
and two naive baselines. This factorial design separates two questions that
single-model studies conflate: whether the \textit{data} contain a
forecastable signal, and whether \textit{machine learning} extracts more of
it than trivial rules do.

\subsection{Models}

Three ensemble algorithms are compared. \textbf{Random Forest}
\citep{breiman2001} averages decorrelated decision trees grown on bootstrap
samples. \textbf{XGBoost} \citep{chen2016} and \textbf{Gradient Boosting
Machines} build trees sequentially, each correcting the residuals of the
last; including both allows comparison of two boosting implementations.
Given the small sample, hyperparameters are fixed at deliberately
conservative values rather than tuned: for XGBoost, maximum depth 2, learning
rate 0.05, 100 estimators, subsample 0.8, and L2 regularization
$\lambda = 1$; the scikit-learn GBM and Random Forest use comparably shallow,
regularized settings. No grid search is performed, because with roughly
13--18 training observations per fold, search procedures select for noise.
As a robustness check on whether tree ensembles are necessary at all, a
\textbf{Ridge regression} on the climate features is evaluated under the
identical protocol.

\subsection{Baselines}

Two naive baselines are evaluated in the same walk-forward loop as the
learned models. \textbf{Persistence} predicts that this year's yield equals
last year's. \textbf{Rolling mean} predicts the mean of the previous three
years. In a series with strong autocorrelation, persistence is a demanding
benchmark; any model that cannot beat it has no forecasting value regardless
of its in-sample fit.

\subsection{Validation: Expanding-Window Walk-Forward}
\label{sec:validation}

Evaluation uses an expanding-window walk-forward design: the models are
trained on 2000--2017 and predict 2018; retrained on 2000--2018 and predict
2019; and so on through 2024. This yields seven genuinely out-of-sample
annual forecasts, each made using only prior information, exactly as an
operational forecasting system would. Random train--test splits are not used
anywhere: with lagged and rolling features, randomly held-out years exchange
information with neighboring training years and the evaluation ceases to
simulate forecasting \citep{bergmeir2012, roberts2017}.

\subsection{Evaluation Metrics}

Performance is summarized by root mean squared error (RMSE) and mean absolute
error (MAE), both in kg/ha, computed over the seven held-out forecasts, with
the coefficient of determination ($R^2$) reported for completeness. With
only seven test points spanning the most volatile period of the series
(including the 2018 collapse and 2021--2022 peak), $R^2$ against the test-set
mean is unstable and frequently negative; RMSE and MAE, and especially their
comparison against the persistence baseline, are the primary metrics. A
per-year error table is reported alongside the aggregates so that no single
year's result is hidden inside an average.

\subsection{Interpretability}

For the best-performing configuration, SHAP values \citep{lundberg2017} are
computed on a model refit to the full series (for interpretation only, after
all evaluation is complete) to identify which features drive predictions and
in which direction.

\section{Results}

\subsection{Agricultural Trends in Sierra Leone, 2000--2024}

Understanding how crop yields and production have moved over the past 25
years provides context for the forecasting results.

\subsubsection{Rice Yield Dynamics}

\begin{figure}[H]
    \centering
    \includegraphics[width=\textwidth]{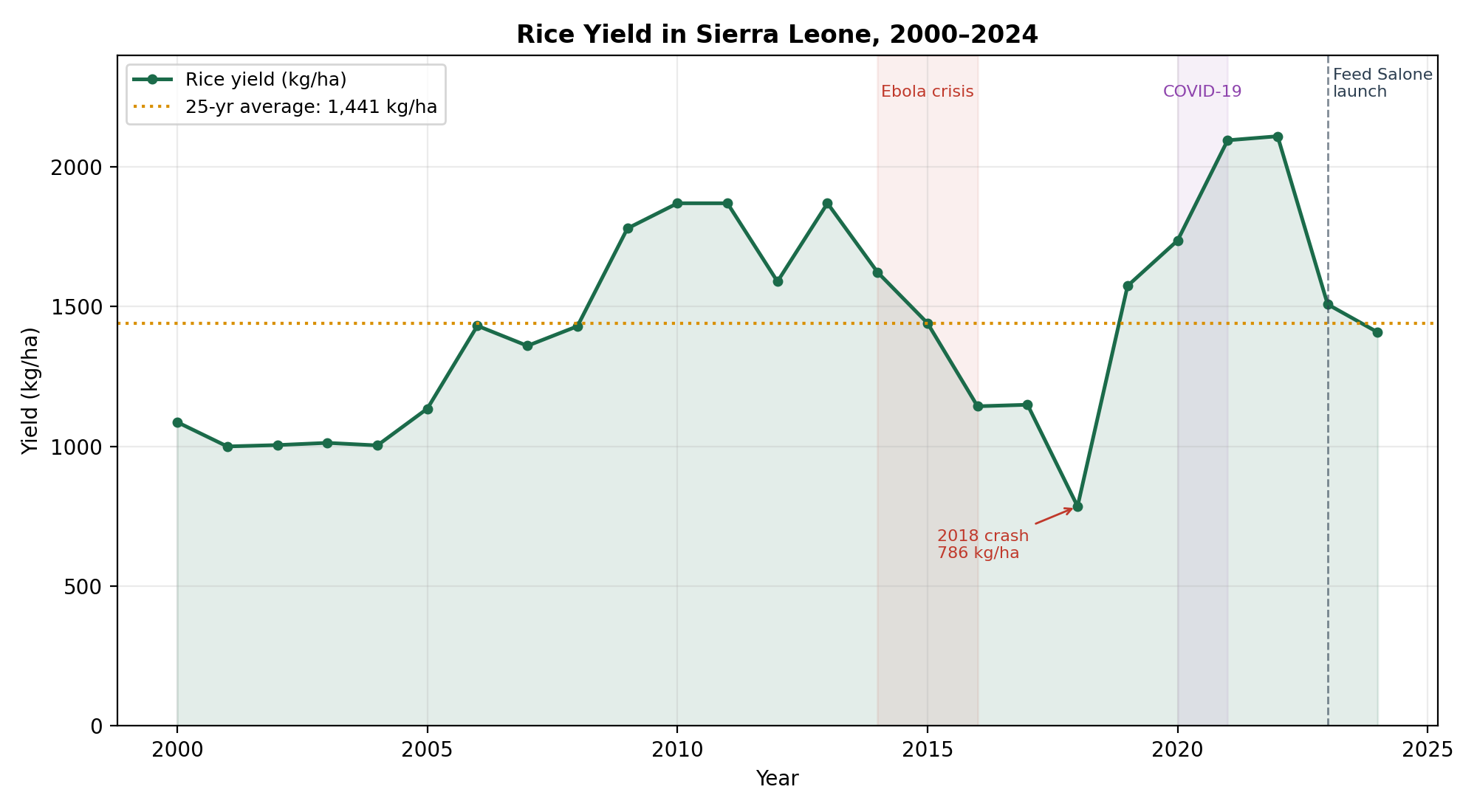}
    \caption{Rice yield in Sierra Leone (2000--2024) with key agricultural
    and economic shock periods marked. The dotted line is the 25-year average
    of 1,441 kg/ha. Source: FAOSTAT (2025).}
    \label{fig:rice_yield_trend}
\end{figure}

Figure~\ref{fig:rice_yield_trend} shows rice yield in Sierra Leone from
2000 to 2024. The 25-year average stands at 1,441 kg/ha, well below the
global average of approximately 4,600 kg/ha and far behind comparable
tropical rice-producing economies such as Thailand ($\sim$3,000 kg/ha) and
Vietnam ($\sim$5,900 kg/ha) \citep{fao2019}.

Three phases characterize the trend. From 2000 to 2005, yields stagnated near
1,000--1,135 kg/ha. From 2006 to 2013 they climbed, reaching 1,780 kg/ha by
2009 and a pre-crisis plateau of 1,870 kg/ha over 2010--2013, likely
reflecting the Smallholder Commercialization Programme and associated input
subsidy interventions \citep{worldbank2022}. The period from 2014 onward has
been defined by volatility and repeated shocks. The Ebola crisis disrupted
agricultural labor supply and rural markets across all producing regions. A
more severe shock followed in 2018, when rice yield collapsed to 786 kg/ha,
the lowest value in the series and a 58 percent decline from the 2013
plateau; lingering post-Ebola disorganization, declining fertilizer access,
and a documented contraction in area harvested converged. Yields then
recovered sharply to a series peak of 2,110 kg/ha in 2022 before contracting
to 1,409 kg/ha in 2024, the first full implementation year of the Feed Salone
Strategy. The forecasting analysis below returns to both the 2018 collapse
and the 2020--2022 recovery, because they set the limits of what any
forecasting model can claim.

\subsubsection{Multi-Crop Yield Patterns}

\begin{figure}[H]
    \centering
    \includegraphics[width=\textwidth]{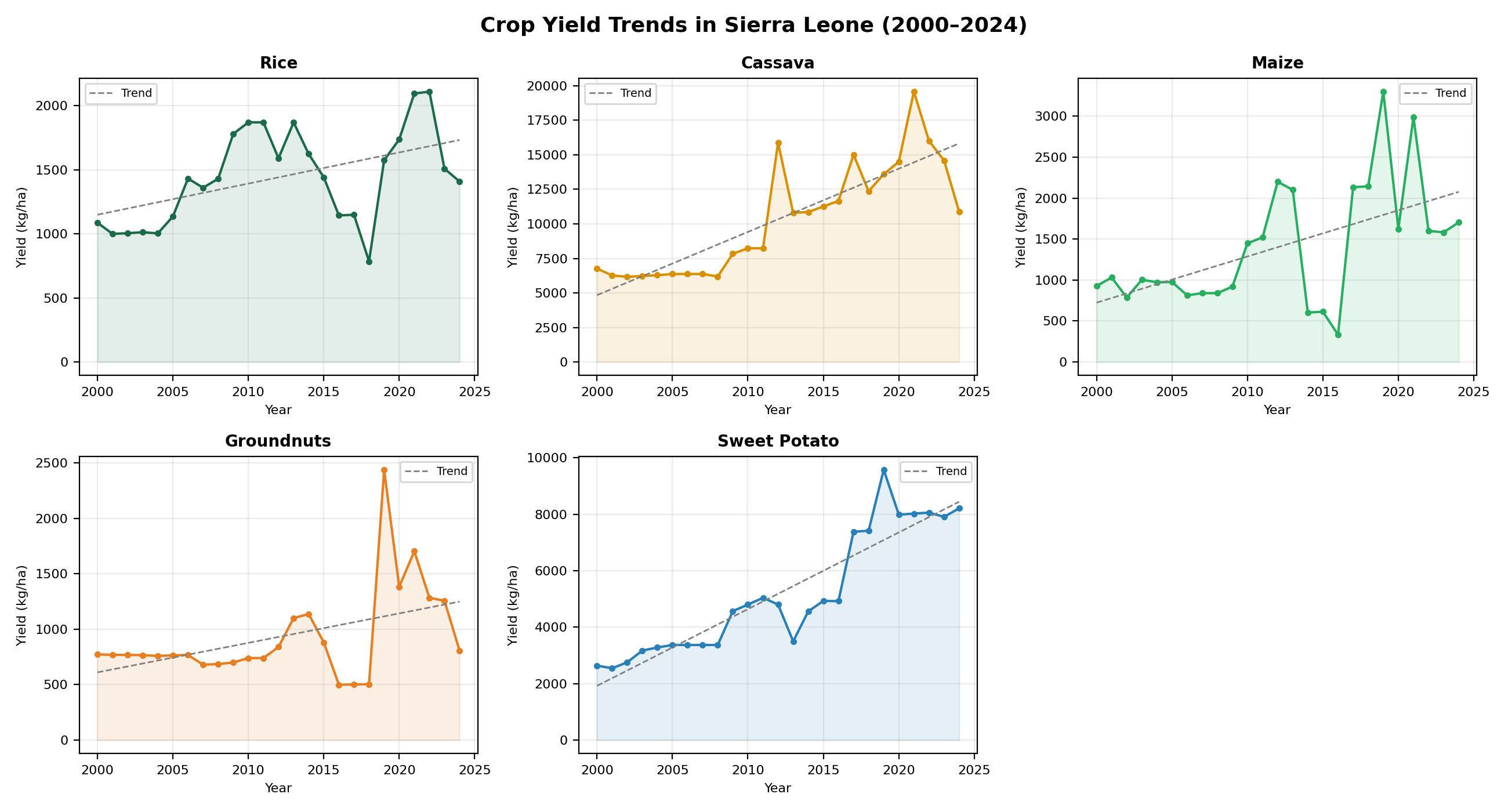}
    \caption{Crop yield trends in Sierra Leone (2000--2024) for key staple
    and cash crops. Dashed lines show linear trend fits. Source: FAOSTAT
    (2025).}
    \label{fig:multi_crop_yields}
\end{figure}

Figure~\ref{fig:multi_crop_yields} extends the picture across five key crops.

Cassava shows the most dramatic long-run improvement, rising from
approximately 6,500 kg/ha in 2000 to a peak of nearly 20,000 kg/ha in 2021.
Sweet potato shows the strongest and most consistent positive trend, growing
from 2,600 kg/ha in 2000 to over 8,000 kg/ha by 2024. Maize and groundnuts
display high year-to-year volatility, with swings of 50 percent or more
between consecutive years, reflecting rainfall sensitivity and the absence of
irrigation in the communities where they are predominantly cultivated. The
groundnut spike to 2,400 kg/ha in 2019 followed by a collapse to 500 kg/ha
the following year is particularly striking and warrants district-level
investigation once microdata become available.

\subsubsection{Production Overview}

\begin{figure}[H]
    \centering
    \includegraphics[width=\textwidth]{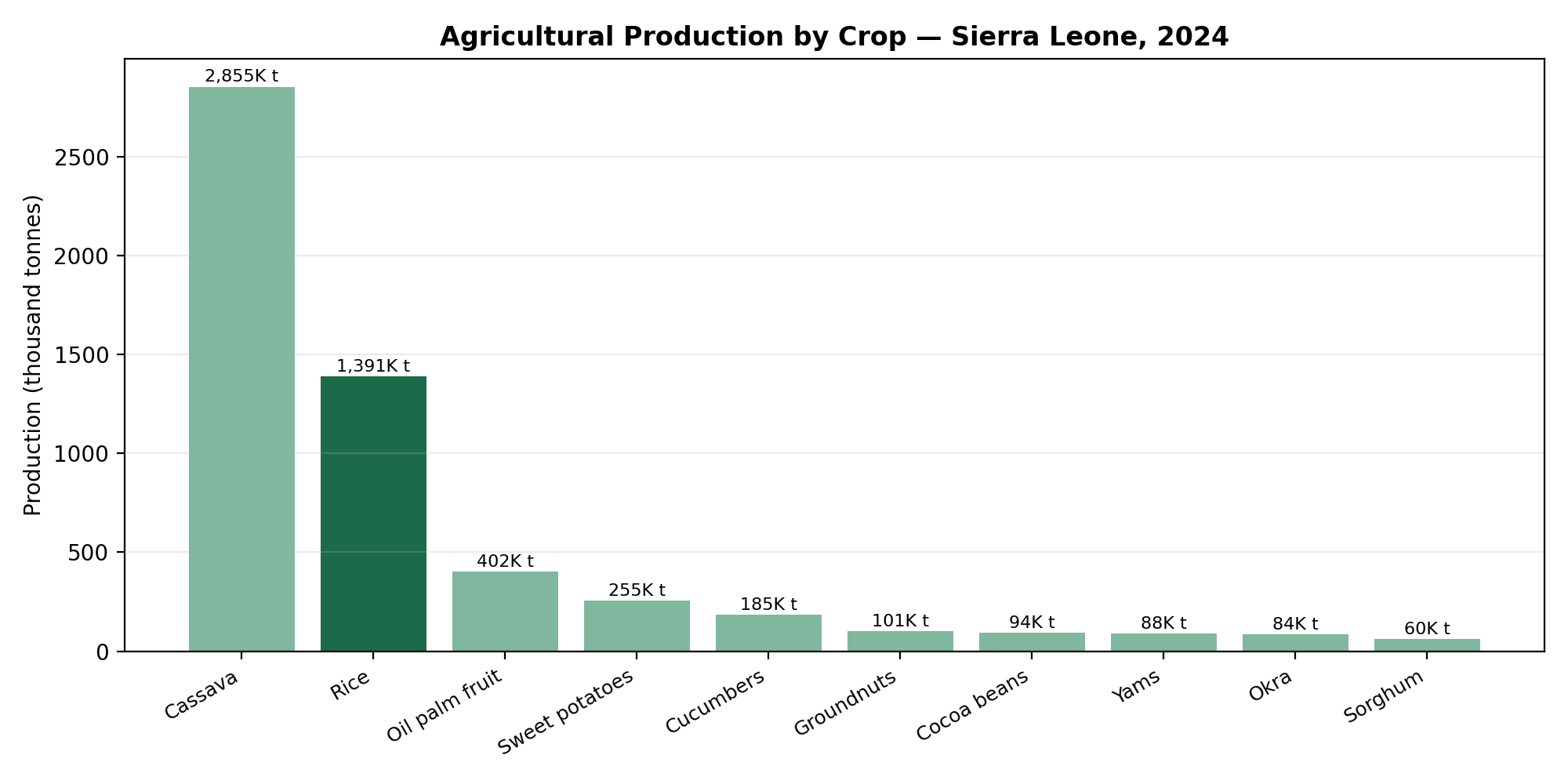}
    \caption{Agricultural production by individual crop, Sierra Leone, 2024.
    Rice, the primary dietary staple, ranks second at 1,391,000 tonnes,
    behind cassava (2,855,000 t). Source: FAOSTAT (2025).}
    \label{fig:production_overview}
\end{figure}

Figure~\ref{fig:production_overview} shows 2024 production for the ten
largest individual crops (FAOSTAT aggregate items are excluded).

Despite being the country's primary dietary staple, rice ranks second in
production volume at 1,391,000 tonnes, less than half of cassava's 2,855,000
tonnes. This imbalance between what Sierra Leone eats and what it grows most
is a direct explanation for the country's persistent rice import dependency,
documented at 480,000 metric tons in 2022, and underscores the urgency of the
rice yield agenda at the center of Feed Salone.

\subsection{Forecasting Performance}
\label{sec:mainresults}

\begin{figure}[H]
    \centering
    \includegraphics[width=\textwidth]{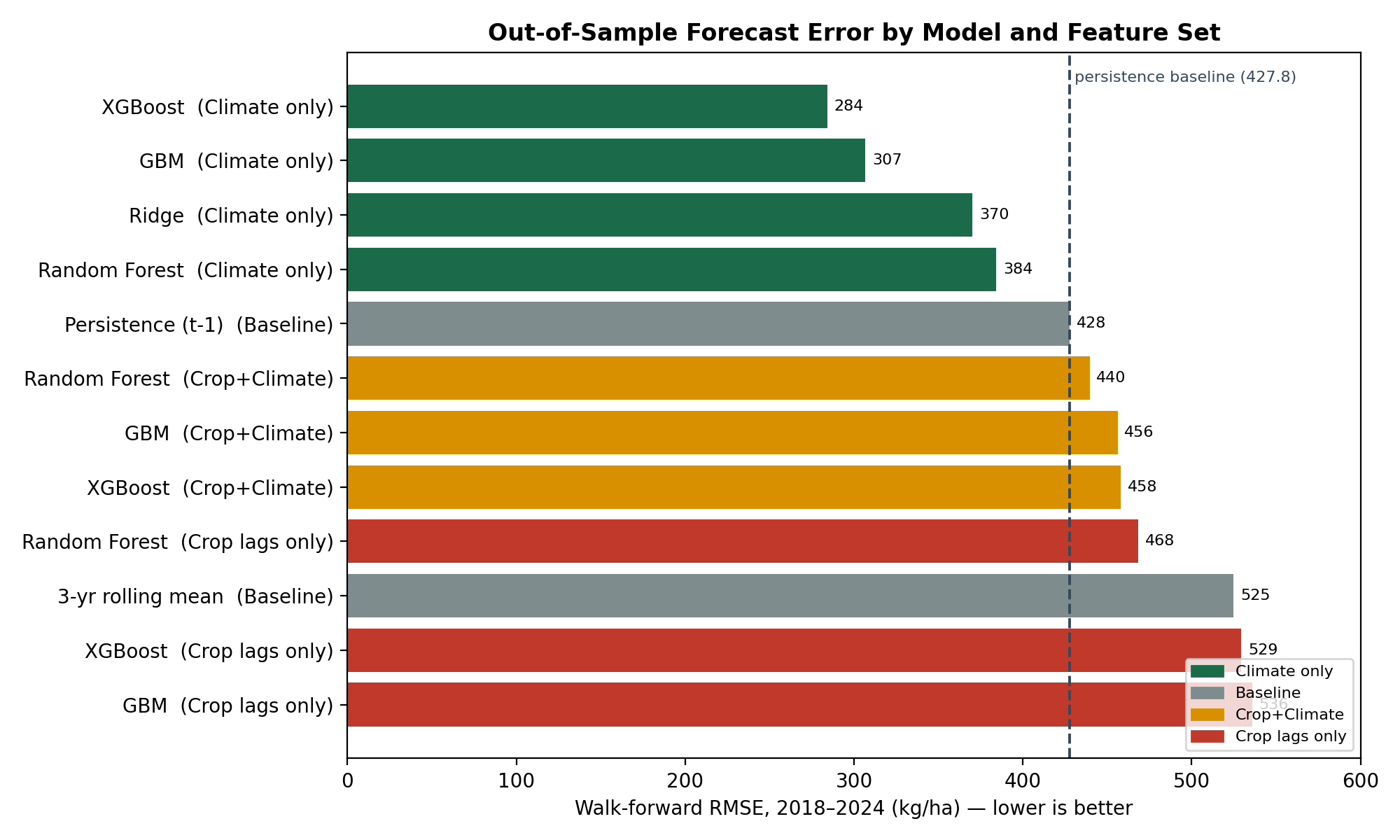}
    \caption{Out-of-sample RMSE by model and feature set, walk-forward
    2018--2024. Only the climate-only configurations beat the persistence
    baseline (dashed line); models given lagged crop statistics, alone or in
    combination, do not.}
    \label{fig:model_comparison}
\end{figure}

Table~\ref{tab:main_results} reports walk-forward forecasting performance for
all model--feature-set combinations and both baselines over the seven
held-out years (2018--2024). Figure~\ref{fig:model_comparison} shows the same
comparison visually.

\begin{table}[H]
\centering
\caption{Walk-forward forecasting performance, rice yield, 2018--2024
(7 out-of-sample years)}
\label{tab:main_results}
\begin{tabular}{llccc}
\toprule
\textbf{Feature set} & \textbf{Model} & \textbf{R\textsuperscript{2}} &
\textbf{RMSE (kg/ha)} & \textbf{MAE (kg/ha)} \\
\midrule
Climate only      & XGBoost            & \textbf{0.542}  & \textbf{284.1} & \textbf{216.3} \\
Climate only      & Gradient Boosting  & 0.465           & 306.8          & 242.7 \\
Climate only      & Ridge (linear)     & 0.222           & 370.2          & 304.0 \\
Climate only      & Random Forest      & 0.162           & 384.2          & 304.6 \\
\midrule
---               & Persistence ($t{-}1$)      & $-0.039$ & 427.8 & 341.3 \\
\midrule
Crop + climate    & Random Forest      & $-0.097$        & 439.5          & 420.5 \\
Crop + climate    & Gradient Boosting  & $-0.181$        & 456.1          & 427.2 \\
Crop + climate    & XGBoost            & $-0.191$        & 458.0          & 427.0 \\
Crop lags only    & Random Forest      & $-0.247$        & 468.5          & 454.7 \\
---               & 3-yr rolling mean  & $-0.565$        & 524.9          & 511.4 \\
Crop lags only    & XGBoost            & $-0.591$        & 529.4          & 497.2 \\
Crop lags only    & Gradient Boosting  & $-0.630$        & 535.8          & 496.7 \\
\bottomrule
\end{tabular}
\vspace{0.5em}
\begin{minipage}{\linewidth}
\small\textit{Expanding-window walk-forward evaluation: train 2000--2017
$\rightarrow$ predict 2018, retrain through 2024. Bold marks the best value
per column. With seven test points spanning the series' most volatile years,
$R^2$ is unstable and frequently negative; RMSE and MAE relative to the
persistence baseline are the primary metrics.}
\end{minipage}
\end{table}

Three results structure everything that follows.

\textbf{First, crop statistics alone cannot forecast rice yield.} Every model
trained on the 35 lagged crop features performs \textit{worse} than naive
persistence (RMSE 468--536 vs.\ 428 kg/ha). The information in last year's
production statistics about this year's yield is already fully captured by
last year's yield itself; the additional crop features contribute noise that
the ensembles, with 13--18 training observations per fold, overfit.

\textbf{Second, free satellite climate data reverses the result.} The
climate-only XGBoost cuts forecast error by one third relative to persistence
(RMSE 284.1 vs.\ 427.8 kg/ha; MAE 216.3 vs.\ 341.3). Critically, this is not
a single-model artifact: every learner improves when given climate features
in place of crop lags, and even the linear Ridge model beats persistence
(RMSE 370.2). The climate signal is real and partly linear; the
XGBoost--Ridge gap of 86 kg/ha suggests additional non-linearity, though with
seven test points that comparison rests heavily on two turning-point years
and is treated as suggestive rather than definitive.

\textbf{Third, more features hurt.} Adding the 35 crop lags \textit{to} the
climate features degrades every model (e.g., XGBoost RMSE rises from 284 to
458). At this sample size, feature parsimony is not a stylistic preference
but a binding statistical constraint, a finding with direct relevance to the
many small-sample agricultural ML studies that maximize feature counts.

\subsection{Where the Gains Come From: Per-Year Analysis}
\label{sec:peryear}

\begin{figure}[H]
    \centering
    \includegraphics[width=\textwidth]{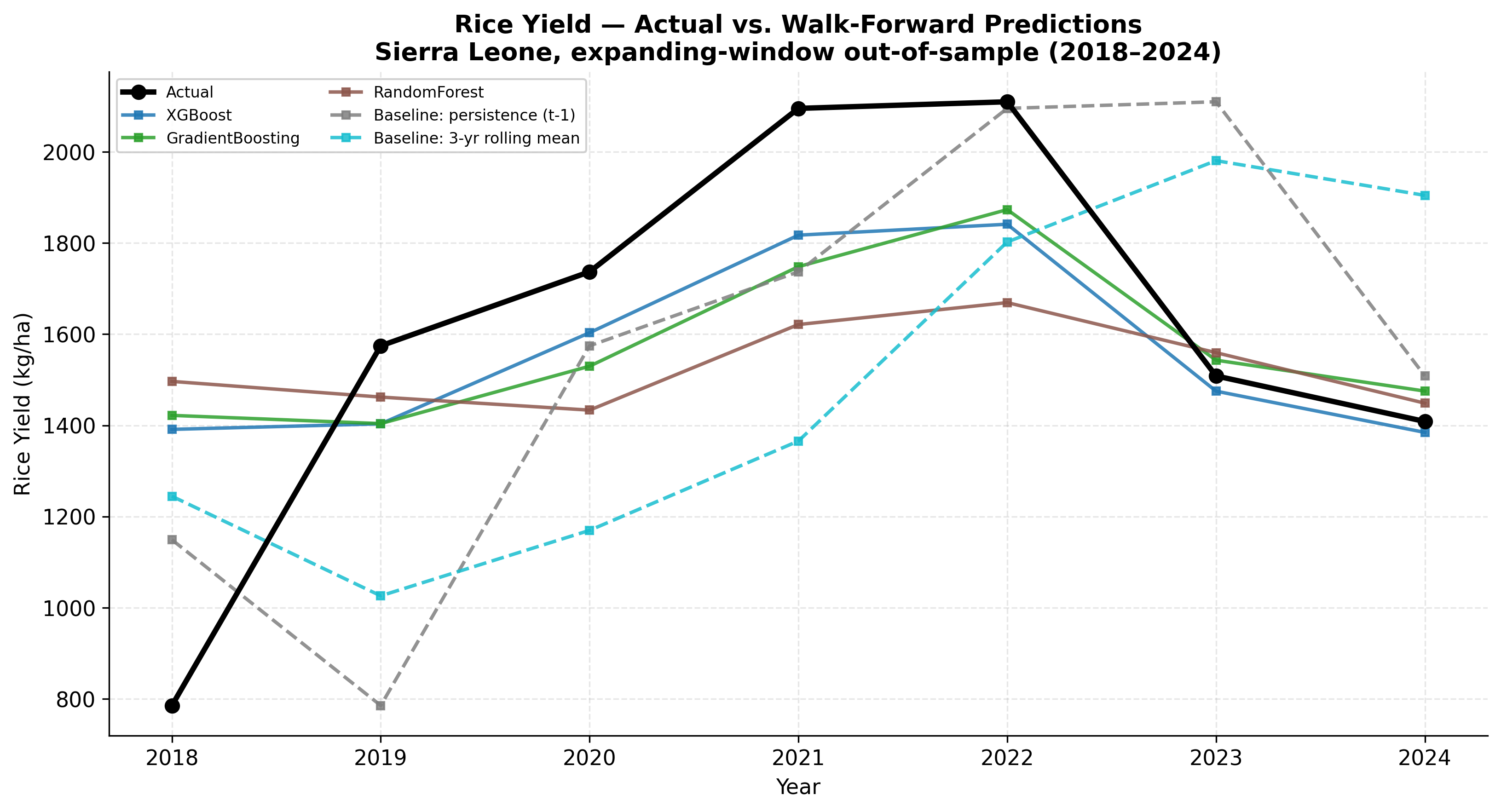}
    \caption{Actual versus walk-forward predicted rice yield by year,
    2018--2024, all models and baselines. The climate-only XGBoost tracks the
    turning points (2019, 2023) that the persistence baseline misses by
    construction; no model anticipates the 2018 collapse.}
    \label{fig:actual_vs_predicted}
\end{figure}

Aggregate error statistics can conceal as much as they reveal with seven test
points, so Table~\ref{tab:per_year} reports the per-year picture for the
decisive years (the full per-year table for all models is archived with the
pipeline). Figure~\ref{fig:actual_vs_predicted} plots actual versus predicted
yield by year.

\begin{table}[H]
\centering
\caption{Per-year absolute forecast error (kg/ha), climate-only XGBoost vs.\
persistence, selected years}
\label{tab:per_year}
\begin{tabular}{lccc}
\toprule
\textbf{Year} & \textbf{Actual yield (kg/ha)} &
\textbf{XGBoost $|$error$|$} & \textbf{Persistence $|$error$|$} \\
\midrule
2018 (collapse)       & 786   & 606          & \textbf{364} \\
2019 (rebound)        & 1{,}574 & \textbf{171} & 789 \\
2023 (contraction)    & 1{,}509 & \textbf{33}  & 601 \\
\bottomrule
\end{tabular}
\vspace{0.5em}
\begin{minipage}{\linewidth}
\small\textit{Bold marks the lower error. The climate model's aggregate
advantage is earned at turning points (2019, 2023), where persistence by
construction lags one year behind. Neither approach anticipated the 2018
collapse. Full table: \texttt{outputs/per\_year\_errors.csv} in the project
repository.}
\end{minipage}
\end{table}

The mechanism behind the climate model's advantage is visible in the table:
persistence, by construction, lags one year behind at every turning point,
while the climate model tracks the yield level. In 2019 the climate model's
error is 171 kg/ha against persistence's 789; in 2023, 33 against 601.
Robustness checks confirm the advantage is not an artifact of any single
year: excluding 2018 entirely \textit{widens} the gap (climate RMSE 182 vs.\
persistence 438 kg/ha).

Equally important is the year both approaches failed. In 2018, when yield
collapsed 32 percent to 786 kg/ha, the climate model predicted 1,391 kg/ha,
an error of 606 kg/ha, worse than persistence. The 2018 collapse was not
climatic in origin; it reflected institutional stressors (post-Ebola
disorganization, fertilizer access, area contraction) invisible to rainfall
and temperature data. The model's value is therefore tracking normal-year
variation, not anticipating structural crises. This boundary is made
explicit here because it determines the appropriate policy use of such a
system (Section~\ref{sec:discussion}).

\subsection{What Drives the Predictions: SHAP Analysis}

\begin{figure}[H]
    \centering
    \includegraphics[width=0.95\textwidth]{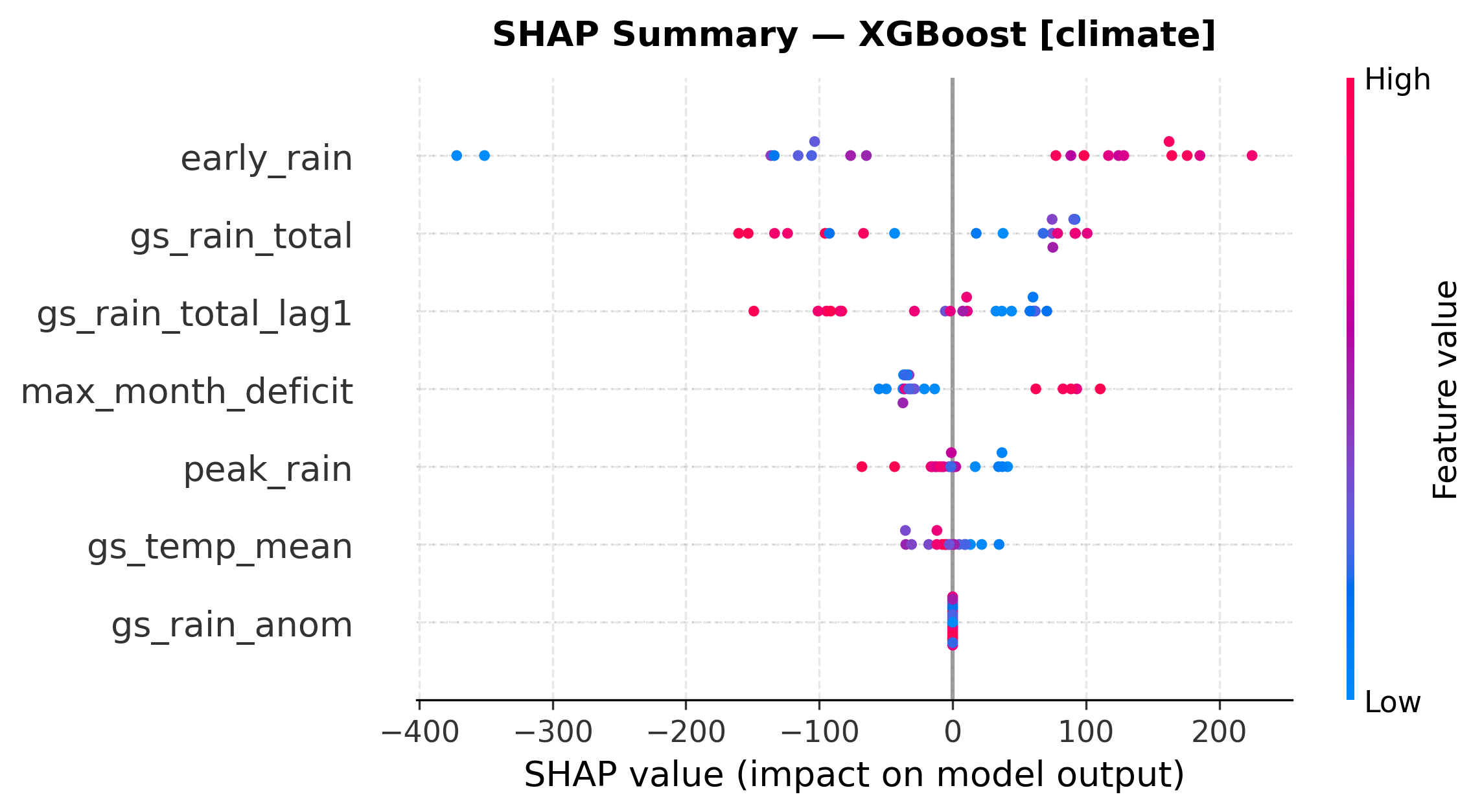}
    \caption{SHAP feature attribution for the climate-only XGBoost model
    (refit on the full series for interpretation after evaluation). Each
    point is one year; the horizontal axis is the feature's contribution to
    predicted yield in kg/ha, colored by feature value (low = blue, high =
    red).}
    \label{fig:shap_summary}
\end{figure}

SHAP attribution identifies early-season rainfall (May--June) as the dominant
driver of the model's predictions, followed by total growing-season rainfall
(May--October) and prior-year growing-season rainfall. The prominence of the
May--June window is the single most policy-relevant pattern in the analysis:
it corresponds to the planting and establishment period for rice in Sierra
Leone's rain-fed systems, and it is fully observable in CHIRPS by early July,
months before harvest. A season's yield risk, to the extent it is climatic,
is therefore knowable in time to act on it.

\begin{figure}[H]
    \centering
    \includegraphics[width=\textwidth]{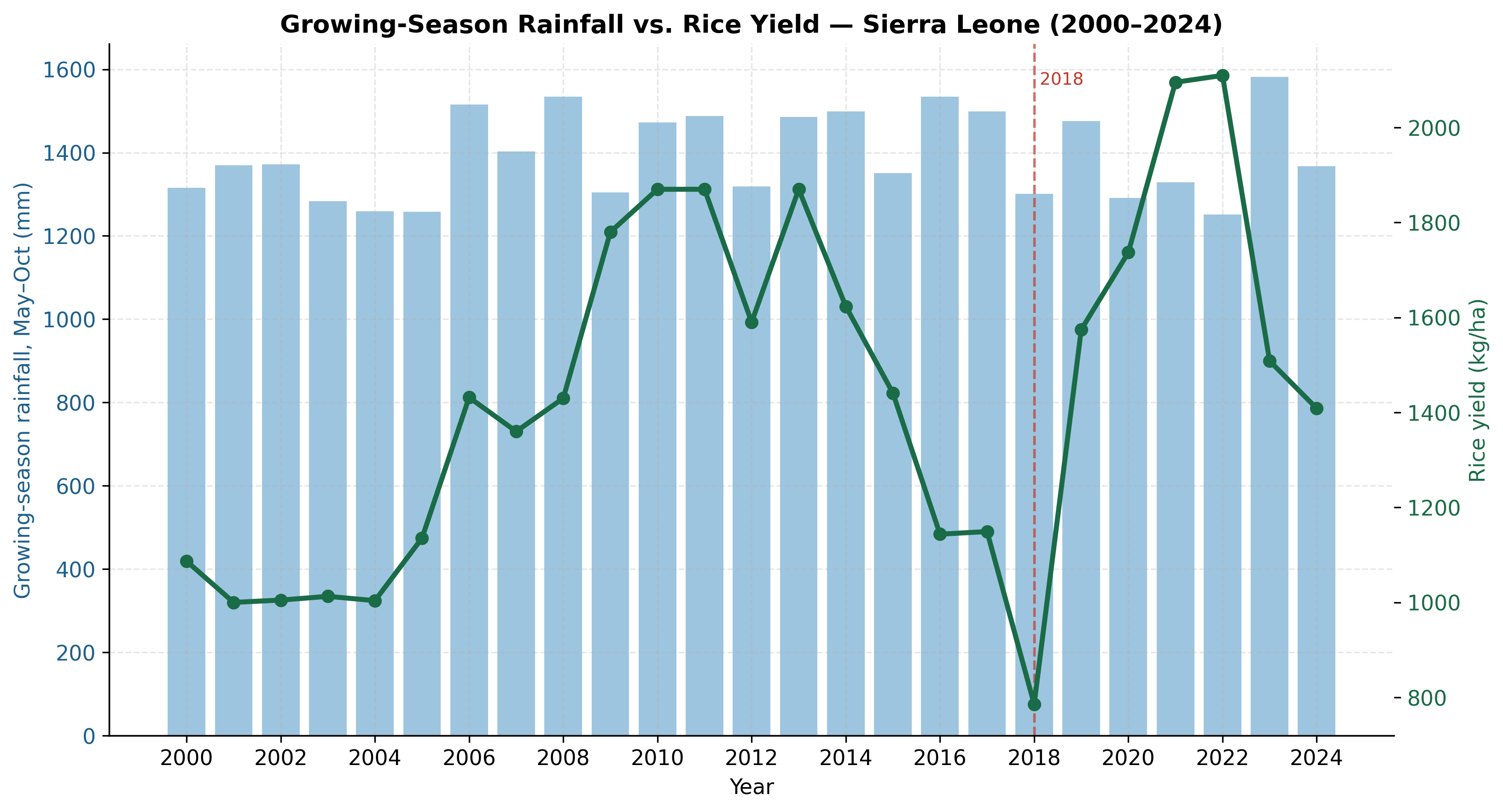}
    \caption{Growing-season rainfall and rice yield by year, 2000--2024.
    The association is positive in typical years; the 2018 collapse and the
    dry-but-high 2020--2022 period mark its limits.}
    \label{fig:rainfall_vs_yield}
\end{figure}

Figure~\ref{fig:rainfall_vs_yield} sets the rainfall--yield relationship
against the full series, including its instructive exceptions: the 2018
collapse occurred without a commensurate rainfall anomaly, and the record
yields of 2020--2022 occurred in below-average rainfall years, consistent
with input-driven gains under the policy programs of that period rather than
favorable weather.

\section{Discussion}
\label{sec:discussion}

\subsection{What the Results Establish, and What They Rule Out}

Read together, the results draw a precise map of what is currently possible.
Sierra Leone's existing crop statistics, used alone, contain no forecastable
signal beyond what last year's yield already provides; a Ministry analyst
equipped only with FAOSTAT extracts cannot build a yield forecasting system,
no matter how sophisticated the algorithm. Adding free satellite climate data
changes that conclusion: forecast error falls by one third, the gain is
robust across model classes and to the exclusion of the most anomalous year,
and the dominant signal, early-season rainfall, is available months before
harvest. At the same time, two boundaries emerge clearly from the analysis.
The model does not anticipate structural collapses whose causes are
institutional rather than climatic, as 2018 demonstrates; and the
dry-but-record-high years of 2020--2022 show that policy and input effects
can dominate weather, which is, from a development perspective, encouraging.

A methodological lesson accompanies the substantive one. An earlier version
of this pipeline, using same-year features, FAOSTAT aggregate items, and a
random 70/30 split, produced an apparent $R^2$ of 0.96. Every component of
that result dissolved under the anti-leakage protocol and walk-forward
evaluation: the aggregate ``cereals'' feature was a near-duplicate of the
target ($r = 0.998$), same-year features answered a question no forecaster
can be asked, and random splitting let test-year information leak through
lagged features. The corrected headline numbers are far smaller and far more
useful. Given how common the flawed design is in the small-sample
agricultural ML literature \citep{kaufman2012, bergmeir2012}, the
before-and-after contrast documented in this study's public repository may
be of independent value to the field.

\subsection{Policy Implications for Feed Salone}

\subsubsection{A Rainfall-Based Early Warning Capability Is Available Now}

The most actionable finding is the dominance of May--June rainfall in the
model's predictions. CHIRPS data for the planting window is published with
short latency and is free; a monitoring workflow that flags anomalously dry
planting seasons by early July would give MAFS several months of lead time
for targeted extension support, input mobilization, and import planning,
before harvest confirms the problem. No new data collection is required to
stand up this capability; it is an analytical product of data that already
exists.

\subsubsection{Forecasting Complements, but Cannot Replace, Institutional Monitoring}

The 2018 lesson cuts the other way: the largest single-year food security
shock in the series was invisible to climate data because its causes were
institutional. A yield intelligence system for Sierra Leone therefore needs
two layers: climate-based seasonal forecasting for weather-driven variation,
and administrative monitoring of input markets, fertilizer availability, and
area planted for the institutional shocks that climate cannot see. The
2020--2022 period offers the optimistic corollary: input-driven gains
overrode unfavorable weather, implying that the policy levers Feed Salone
controls are powerful enough to dominate climate variation in normal years.

\subsubsection{The Data-Infrastructure Case, Restated Empirically}

This study's central institutional finding is no longer rhetorical. The
country's official agricultural statistics are thin: only 467 of 2,593
FAOSTAT records for Sierra Leone are flagged as official figures, and the
most recent rice yields are FAO estimates. That thinness has a measurable
cost: the statistics cannot support forecasting. The marginal value of
better data is equally measurable: one free satellite product moved the
system from useless to useful. District-level microdata such as the 2023
SLAASS is the next increment, and the one that would move prediction to the
administrative level where extension decisions are actually made. MAFS's
PEMSD division currently lacks the technical capacity to operate such
analytics \citep{fidinnovation2024}; the entire pipeline used here is openly
archived to lower that barrier.

\subsection{Limitations}
\label{sec:limitations}

Five limitations should be acknowledged. First, national aggregation masks
the district-level heterogeneity in soils, rainfall, and practices that
matters most for targeting; only microdata can relax this. Second, the
evaluation rests on seven out-of-sample years, the maximum the series allows
under honest walk-forward design, but few enough that all aggregate metrics,
and especially the XGBoost--Ridge comparison, should be read as indicative.
Third, roughly half the FAOSTAT records are FAO estimates rather than
official figures, including the 2022--2024 rice yields; measurement error in
the outcome bounds achievable accuracy. Fourth, the climate features are
national averages of products with their own uncertainties, partially
mitigated by the CHIRPS--NASA POWER cross-check ($r = 0.73$). Fifth,
post-harvest loss modeling, a co-equal objective of this research agenda,
requires household-level data on storage, transport, and handling that exist
in the 2023 SLAASS but were not accessible during the study period despite
registration on the FAO microdata portal; it is the immediate next step of
this work rather than a delivered result.

\section{Conclusion and Policy Recommendations}

This paper set out to answer a question that Sierra Leone's Feed Salone
Strategy implicitly assumes has a positive answer: can the country's
agricultural outcomes be forecast from available data? The answer is
conditional, and the conditions themselves are informative. Twenty-five
years of national crop statistics, evaluated under a strict anti-leakage
protocol and walk-forward validation, cannot outperform the naive rule that
next year will resemble this one. Free satellite climate data, CHIRPS
rainfall above all, changes the answer: forecast error falls by one third,
the gain holds across model classes, and the dominant predictor, May--June
planting-season rainfall, is observable months before harvest. The
boundaries are equally clear: institutional shocks like the 2018 collapse are
invisible to climate data, and input-driven policy effects can dominate
weather, as 2020--2022 showed.

Five policy recommendations follow from the empirical findings.

\bigskip
\noindent\textbf{Recommendation 1: Stand up a CHIRPS-based seasonal rainfall
monitor within MAFS.}
The single strongest predictor of rice yield, May--June rainfall, is free,
published with short latency, and observable by early July. A
planting-season rainfall dashboard flagging anomalously dry starts would give
MAFS months of lead time for extension support and import planning, at
essentially zero data cost.

\bigskip
\noindent\textbf{Recommendation 2: Build a two-layer yield intelligence
system.}
Pair climate-based seasonal forecasting with administrative monitoring of
input markets, fertilizer availability, and area planted. The 2018 collapse
demonstrates that the largest shocks can be institutional, not climatic, and
no climate model will see them coming.

\bigskip
\noindent\textbf{Recommendation 3: Front-load post-harvest infrastructure
investment.}
Cassava, the country's largest crop by volume at 2.9 million tonnes, is also
its most spoilage-prone \citep{sheahan2017}. If Feed Salone's productivity
agenda succeeds, bumper harvests will strain storage and processing capacity
precisely when output is highest; that capacity should be built before the
gains arrive, not in response to them.

\bigskip
\noindent\textbf{Recommendation 4: Prioritize SLAASS microdata access and
district-level modeling.}
National aggregates are a binding ceiling on what forecasting can deliver.
The 2023 SLAASS microdata, covering 5,200 households across 520 enumeration
areas, would extend prediction and post-harvest loss modeling to the district
level where extension decisions are made. Streamlining researcher access to
this data is among the cheapest high-return actions available to MAFS and
Statistics Sierra Leone.

\bigskip
\noindent\textbf{Recommendation 5: Invest in Sierra Leonean data science
capacity.}
Every result in this paper was produced with free data and open-source tools;
the constraint is not data or technology but analytical capacity within the
country's institutions \citep{fidinnovation2024}. The most durable path to
data-driven agricultural governance is a generation of Sierra Leonean
researchers and analysts who understand both the tools and the context.
Programs like RiseAfrica Foundation's RiseLab represent exactly that
investment, and the full pipeline behind this study is archived publicly to
support it.

\bigskip\bigskip

\noindent Agriculture has always been how Sierra Leone has fed itself. This
study shows precisely where data science can now help, and where it cannot
yet. The rainfall signal is free and waiting to be used. The deeper gains
wait on the data the country has already collected but not yet opened, and on
the people trained to use it. What remains is the institutional commitment to
bring them together.

\section*{Data Availability Statement}

All data used in this study are publicly available. FAOSTAT crop statistics
for Sierra Leone (2000--2024) are available from
\url{https://www.fao.org/faostat}. CHIRPS v2.0 monthly precipitation rasters
are available from the Climate Hazards Center, University of California,
Santa Barbara (\url{https://www.chc.ucsb.edu/data/chirps}). NASA POWER
agroclimatology data are available from
\url{https://power.larc.nasa.gov}. Administrative boundaries are from
geoBoundaries (\url{https://www.geoboundaries.org}). The complete analytical
pipeline, including all preprocessing, feature engineering, model training,
evaluation code, derived datasets, and per-year results, is archived at
\url{https://github.com/Denis060/sierraleone-agri-ml}.

\section*{Conflict of Interest Statement}

The author declares no conflict of interest. The author is the founder of the
RiseAfrica Foundation for STEM and Innovation, a non-profit organization
referenced in Recommendation 5; this affiliation is disclosed on the title
page.

\section*{Acknowledgments}

The author thanks the Food and Agriculture Organization of the United
Nations, the Climate Hazards Center at UC Santa Barbara, and the NASA POWER
project for maintaining the open data resources on which this study depends.

\newpage
\bibliographystyle{apalike}

\begin{thebibliography}{99}

\bibitem{worldbank2022}
World Bank. (2022). \textit{Agricultural Data and Smallholder Decision-Making
in Sub-Saharan Africa}. Washington, DC: World Bank Group.

\bibitem{usda2023}
United States Department of Agriculture. (2023). \textit{Sierra Leone:
Agriculture Sector Country Commercial Guide}. Washington, DC: USDA Foreign
Agricultural Service.

\bibitem{statssl2023}
Statistics Sierra Leone \& Ministry of Agriculture and Food Security. (2023).
\textit{2023 Sierra Leone Annual Agricultural Sample Survey Report}. Freetown:
Stats SL/MAFS. Available at: \url{https://www.statistics.sl}

\bibitem{feedsalone2023}
Government of Sierra Leone. (2023). \textit{Feed Salone Strategy 2023--2030:
A Blueprint for Agricultural Transformation in Sierra Leone}. Freetown:
Office of the President / Ministry of Agriculture and Food Security.

\bibitem{fidinnovation2024}
Fund for Innovation in Development. (2024). \textit{Building Data-Driven
Agricultural Policy in Sierra Leone}. Project Documentation. Available at:
\url{https://fundinnovation.dev}

\bibitem{50x2030_2024}
50x2030 Initiative. (2024). \textit{One Year of Feed Salone: Progress Toward
Food Security Objectives in Sierra Leone with 50x2030 Support}. Rome: FAO/IFAD.

\bibitem{affognon2015}
Affognon, H., Mutungi, C., Sanginga, P., \& Borgemeister, C. (2015).
Unpacking post-harvest losses in sub-Saharan Africa: A meta-analysis.
\textit{World Development}, 66, 49--68.

\bibitem{fao2019}
Food and Agriculture Organization of the United Nations. (2019).
\textit{The State of Food and Agriculture: Moving Forward on Food Loss and
Waste Reduction}. Rome: FAO.

\bibitem{nrds2022}
Government of Sierra Leone. (2022). \textit{National Rice Development
Strategy II (NRDS2)}. Freetown: Ministry of Agriculture and Food Security.

\bibitem{worldbank2023}
World Bank. (2023). \textit{Sierra Leone Overview: Development News, Research,
Data}. Washington, DC: World Bank Group.

\bibitem{wdi2024}
World Bank. (2024). \textit{World Development Indicators: Agriculture,
forestry, and fishing, value added (\% of GDP) -- Sierra Leone}. Washington,
DC: World Bank Group. Available at: \url{https://data.worldbank.org}

\bibitem{frontiers2026}
Frontiers in Artificial Intelligence. (2026). Maize yield prediction using
machine learning: A systematic literature review.
\textit{Frontiers in Artificial Intelligence}, 9.
\url{https://doi.org/10.3389/frai.2026.1735157}

\bibitem{tandason2023}
Tandason, A.F., Bah, A.S., George, D.R., Musa, M.F., \& Sheriff, H.M. (2023).
Assessing post-harvest losses of rice processing at Agricultural Business
Centers (ABCs) in rice production in Sierra Leone.
\textit{International Journal of Agricultural Economics}, 8(2), 48--67.

\bibitem{sciencedirect2022}
Crops yield prediction based on machine learning models: Case of West African
countries. (2022). \textit{Smart Agricultural Technology}, 2, 100168.

\bibitem{vanklompenburg2020}
Van Klompenburg, T., Kassahun, A., \& Catal, C. (2020). Crop yield prediction
using machine learning: A systematic literature review.
\textit{Computers and Electronics in Agriculture}, 177, 105709.

\bibitem{liakos2018}
Liakos, K.G., Busato, P., Moshou, D., Pearson, S., \& Bochtis, D. (2018).
Machine learning in agriculture: A review. \textit{Sensors}, 18(8), 2674.

\bibitem{arrieta2020}
Arrieta, A.B., D\'{i}az-Rodr\'{i}guez, N., Del Ser, J., et al. (2020).
Explainable Artificial Intelligence (XAI): Concepts, taxonomies, opportunities
and challenges toward responsible AI.
\textit{Information Fusion}, 58, 82--115.

\bibitem{sheahan2017}
Sheahan, M., \& Barrett, C.B. (2017). Review: Food loss and waste in
sub-Saharan Africa. \textit{Food Policy}, 70, 1--12.

\bibitem{breiman2001}
Breiman, L. (2001). Random forests. \textit{Machine Learning}, 45(1), 5--32.

\bibitem{chen2016}
Chen, T., \& Guestrin, C. (2016). XGBoost: A scalable tree boosting system.
\textit{Proceedings of the 22nd ACM SIGKDD International Conference on
Knowledge Discovery and Data Mining}, 785--794.

\bibitem{lundberg2017}
Lundberg, S.M., \& Lee, S.I. (2017). A unified approach to interpreting model
predictions. \textit{Advances in Neural Information Processing Systems}, 30.

\bibitem{lobell2010}
Lobell, D.B., \& Burke, M.B. (2010). On the use of statistical models to
predict crop yield responses to climate change.
\textit{Agricultural and Forest Meteorology}, 150(11), 1443--1452.

\bibitem{kaufman2012}
Kaufman, S., Rosset, S., Perlich, C., \& Stitelman, O. (2012). Leakage in
data mining: Formulation, detection, and avoidance.
\textit{ACM Transactions on Knowledge Discovery from Data}, 6(4), 1--21.

\bibitem{bergmeir2012}
Bergmeir, C., \& Ben\'{i}tez, J.M. (2012). On the use of cross-validation for
time series predictor evaluation. \textit{Information Sciences}, 191,
192--213.

\bibitem{roberts2017}
Roberts, D.R., Bahn, V., Ciuti, S., et al. (2017). Cross-validation
strategies for data with temporal, spatial, hierarchical, or phylogenetic
structure. \textit{Ecography}, 40(8), 913--929.

\bibitem{funk2015}
Funk, C., Peterson, P., Landsfeld, M., et al. (2015). The climate hazards
infrared precipitation with stations --- a new environmental record for
monitoring extremes. \textit{Scientific Data}, 2, 150066.

\bibitem{runfola2020}
Runfola, D., Anderson, A., Baier, H., et al. (2020). geoBoundaries: A global
database of political administrative boundaries. \textit{PLoS ONE}, 15(4),
e0231866.

\end{thebibliography}

\end{document}